\documentclass[letterpaper, 10 pt, conference]{classes/ieeeconf}  
\IEEEoverridecommandlockouts 
\usepackage{subcaption}
\usepackage{packages/decar-common}
\usepackage{packages/decar-dynamics}
\usepackage[backend=bibtex,bibstyle=ieee,citestyle=numeric-comp,doi=false,isbn=false]{biblatex}
\usepackage{graphicx}
\usepackage{booktabs}
\usepackage{lipsum}
\usepackage{derivative}
\usepackage{soul}
\usepackage{tikz}
\usepackage{todonotes}
\usepackage{algorithm}
\usepackage{algpseudocode}

\usepackage[hidelinks]{hyperref}

\DefineBibliographyStrings{english}{%
  andothers = {\textit{et\addabbrvspace al\adddot}},
}

\bibliography{refs}
\graphicspath{{figs/}}

\title{\LARGE \bf
    ARC: Adaptive Robust Joint State and Covariance Estimation
}

\author{Alexandre Hadji-Thomas, Andrew Stirling, and James R. Forbes}

\begin{document}
    \maketitle
    \thispagestyle{empty}
    \pagestyle{empty}
    
    \begin{abstract}
Sensor measurements are frequently corrupted by outliers and 
non-Gaussian noise. These imperfections in the sensor data can 
cause classical state estimators to generate biased and 
unreliable state and uncertainty estimates. Robust estimators reject or downweight outliers but do not perform measurement covariance estimation, whereas joint 
state and covariance estimators assume Gaussian residuals and fixed loss shape parameters. 
Integrating these two capabilities into a single framework is an opportunity to simultaneously estimate both state and covariance in the presence of outliers. This paper proposes a unified Block-Coordinate 
Descent framework that combines a norm-aware adaptive robust 
loss, an Iteratively Reweighted Least-Squares state update, and 
a Minimum Weighted Covariance Determinant covariance estimator, 
yielding a self-tuning joint state and covariance estimator. 
The framework is evaluated in a 
Monte-Carlo simulation and on real-world ultra-wideband 
localization experiments in cluttered non-line-of-sight 
environments. Results show that the proposed estimator consistently 
recovers the true inlier measurement covariance and matches 
or exceeds the state estimation accuracy of all baselines, 
without requiring any manual parameter tuning.
\end{abstract}
    \section{Introduction}
Reliable navigation is essential to autonomous robot operation. 
A standard means of estimating a robot's state is by minimizing 
a weighted least-squares (LS) cost function using residuals between 
measurements and a model that is a function of the robot state. 
This formulation assumes that the noise is Gaussian and that the 
measurement covariance is known and fixed~\cite{Barfoot_2025}. However, these 
assumptions may not be satisfied in practice. In particular, when 
measurements are corrupted by outliers, the state estimate becomes 
biased and the covariance predictions unreliable~\cite{RN12}.

For example, when using ultra-wideband (UWB) radio for localization 
indoors, obstacles and clutter frequently create 
non-line-of-sight (NLOS) conditions that introduce heavy-tailed 
measurement errors and positive range biases~\cite{Guvenc2009NLOS}. 
These non-Gaussian effects can degrade the accuracy of the state 
estimate and the reliability of the associated uncertainty 
provided by the weighted LS solution.

\usetikzlibrary{shapes.geometric, arrows.meta, positioning, calc}

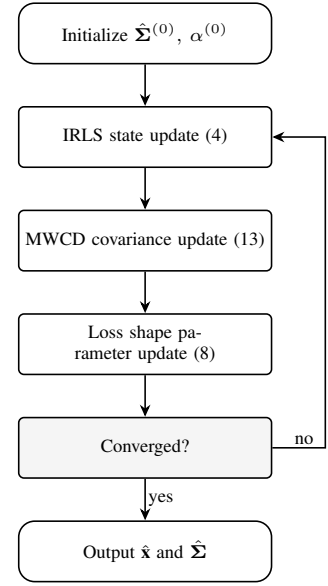
\begin{figure}[htbp!]
\centering
\begin{tikzpicture}[
    node distance=0.55cm,
    every node/.style={inner sep=2pt},
    proc/.style={rectangle, rounded corners=2pt, draw, semithick, align=center,
                 text width=3.2cm, minimum height=0.8cm, font=\scriptsize},
    dec/.style={rectangle, rounded corners=2pt, draw, semithick, align=center,
                fill=black!4, text width=3.2cm, minimum height=0.8cm, font=\scriptsize},
    term/.style={rectangle, rounded corners=6pt, draw, semithick, align=center,
                 text width=3.2cm, minimum height=0.8cm, font=\scriptsize},
    arr/.style={-{Stealth[length=1.6mm,width=1.4mm]}, semithick},
    lbl/.style={font=\scriptsize, inner sep=1pt}
]

\node[term] (init) {Initialize $\hat{\mbs{\Sigma}}^{(0)},\, \alpha^{(0)}$};
\node[proc, below=of init] (state) {IRLS state update~\eqref{eq:robust_loss}};
\node[proc, below=of state] (cov) {MWCD covariance update~\eqref{eq:mwcd_covariance}};
\node[proc, below=of cov] (alpha) {Loss shape parameter update~\eqref{eq:alpha_optimization}};
\node[dec, below=of alpha] (check) {Converged?};
\node[term, below=of check] (done) {Output $\hat{\mbf{x}}$ and $\hat{\mbs{\Sigma}}$};

\draw[arr] (init) -- (state);
\draw[arr] (state) -- (cov);
\draw[arr] (cov) -- (alpha);
\draw[arr] (alpha) -- (check);
\draw[arr] (check) -- node[lbl, right] {yes} (done);

\draw[arr] (check.east) -- ++(0.7,0)
    node[lbl, above, pos=0.6] {no}
    |- (state.east);

\end{tikzpicture}
\caption{Flowchart of the proposed adaptive robust joint state and covariance
estimation procedure.}
\label{fig:algorithm_flow}
\end{figure}
\
\subsection{Related Work}
Robust state estimation methods have been developed to mitigate the 
influence of outliers in LS problems. \textcite{Huber1964} 
introduced M-estimators with bounded influence functions, and 
\textcite{BlackRangarajan1996} unified robust 
statistics with outlier rejection through the duality between 
robust loss functions and latent-variable reweighting schemes. 
\textcite{7158322} compared several fixed robust cost functions 
in an iteratively reweighted least squares (IRLS) scheme and found that the error deflation term, or other 
equivalent tuning parameter, must be carefully selected, especially 
in the presence of poor initial conditions, highlighting the cost of 
relying on a hand-tuned, fixed loss form.
\textcite{RN21, 9361339} proposed a general adaptive loss function 
that subsumes many existing robust losses and automatically 
adapts its loss shape parameter to the residual distribution, 
eliminating the need for manual selection. This was later 
extended by \textcite{RN20} to multivariate 
LS problems, where the Mahalanobis distance residual 
follows a Chi distribution with a non-zero mode, introducing 
a norm-aware weighting scheme that correctly identifies inliers 
regardless of the error dimension. However, none of these methods 
provide a mechanism for estimating or adapting the measurement 
covariance, which are assumed to be fixed and known.

Joint state and covariance estimation methods address this
limitation by treating the measurement covariance as an unknown
variable estimated alongside the state. \textcite[Sec.~5.5.3]{Barfoot_2025}
reviews methods that alternate between state
and covariance updates, showing that the joint MAP covariance
estimation problem admits a convex structure. Robustness is
incorporated through M-estimators with IRLS to jointly compute
the covariance and state estimates, but the loss shape parameter
is not adapted to the empirical residual distribution.
\textcite{Pfeifer2023AdaptiveEstimation} proposes 
an adaptive estimation framework using Gaussian mixture models 
to handle non-Gaussian noise distributions, estimating the state
alongside a mixture uncertainty model, but does not 
explicitly recover the inlier measurement covariance and requires 
the number of mixture components to be specified a priori. None 
of these methods jointly adapt the loss shape parameter and the 
measurement covariance within a unified framework, leaving them 
unable to automatically tune their robustness level in response 
to changing noise conditions.

Robust covariance estimation methods offer an alternative approach 
to handling outlier-contaminated measurements. The Minimum 
Covariance Determinant (MCD) estimator~\cite{RN22} identifies 
the subset of observations whose empirical covariance has the 
smallest determinant, achieving a breakdown point of up to 50\%, 
where the breakdown point is the maximum fraction of
outliers an estimator can tolerate before producing an arbitrarily
biased result.
Kalina and Tichavský~\cite{Kalina2022WeightedMCD} extended this 
to the Minimum Weighted Covariance Determinant (MWCD) estimator, 
incorporating continuous residual weights to improve covariance 
reestimation accuracy. However, neither method has been integrated with an adaptive 
robust state estimator for joint state and covariance estimation 
under outlier-contaminated measurements.

None of these approaches offer a unified mechanism for jointly 
adapting the loss shape parameter and recovering the measurement 
covariance while estimating the state, leaving a gap in the 
literature for a self-tuning robust joint state and covariance estimator capable of 
handling outlier-contaminated and unknown noise statistics within 
a single optimization framework.

\subsection{Contributions and Paper Organization}
The contributions of this work are as follows.
\begin{itemize}
    \item Proposing a unified Block-Coordinate Descent (BCD) framework that combines a norm-aware 
    adaptive robust loss, an IRLS state update, and a MWCD covariance 
    estimator for joint state and covariance estimation under unknown 
    and outlier-contaminated measurement noise.
    \item Demonstrating that jointly adapting the loss shape parameter 
    and the measurement covariance within a single BCD cycle yields 
    a self-tuning estimator that requires no manual parameter tuning.
    \item Validating the framework on real-world UWB localization 
    experiments in cluttered NLOS environments, demonstrating 
    consistent recovery of the true inlier measurement covariance 
    and better state estimation accuracy than all baselines, 
    without any manual parameter tuning.
\end{itemize}

The remainder of this paper is organized as follows. 
Section~\ref{sec:preliminaries} reviews the mathematical 
background. Section~\ref{sec:adaptive_robust} presents the proposed 
adaptive robust joint estimation framework. 
Section~\ref{sec:results} validates the method on real-world 
UWB localization experiments. Section~\ref{sec:discussion} 
discusses the results. Section~\ref{sec:conclusion} concludes 
the paper.
    \section{Preliminaries}
\label{sec:preliminaries}


\subsection{Joint State and Covariance 
    Estimation}
\label{sec:joint}
Joint state and covariance estimation seeks to simultaneously 
estimate the system state and the measurement noise covariance 
by solving the optimization problem
\begin{equation}
    \label{eq:joint_state_covariance_estimation}
    \{\hat{\mbf{x}},\, \hat{\mbs{\Sigma}}\}
    = \arg\min_{\mbf{x},\, \mbs{\Sigma}} 
    \mathcal{J}(\mbf{x},\, \mbs{\Sigma}),
\end{equation}
where $\mbf{x} \in \mathbb{R}^{(N \times n_\mathrm{x}) \times 1}$ is the stacked 
state vector, $\mbf{x} = [\mbf{x}_1^\trans \; \mbf{x}_2^\trans \; \cdots \; \mbf{x}_N^\trans]^\trans$, over all $N$ time steps and $n_\mathrm{x}$ is the 
state dimension, such that $\mbf{x}_\mathrm{i} \in \mathbb{R}^{n_\mathrm{x}}$.
The measurement error is defined as 
$\mbf{e}_\mathrm{i} = \mbf{y}_\mathrm{i} - \mbf{g}_\mathrm{i}(\mbf{x}_\mathrm{i})$, 
where $\mbf{y}_\mathrm{i}$ is the $\mathrm{i}$-th noisy measurement and 
$\mbf{g}_\mathrm{i}(\mbf{x}_\mathrm{i})$ is the measurement 
function evaluated at the current state estimate, and 
$\mbf{e}_\mathrm{i} \sim \mathcal{N}(\mbf{0}, \mbs{\Sigma})$ 
with $\mbs{\Sigma}$ the measurement noise covariance matrix. The associated 
scalar residual is the Mahalanobis distance,
\begin{equation}
    \epsilon_\mathrm{i} 
    = \left\Vert 
        \mbf{e}_\mathrm{i}
    \right\Vert_{\mbs{\Sigma}^{-1}} 
    = \sqrt{\mbf{e}_\mathrm{i}^\trans \mbs{\Sigma}^{-1} \mbf{e}_\mathrm{i}}.
    \label{eq:mahalanobis}
\end{equation}
The joint 
negative log-likelihood cost, which takes the form of a weighted 
least-squares objective, is given by
\begin{equation}
    \mathcal{J}(\mbf{x}, \mbs{\Sigma}) = 
    \underbrace{\frac{N}{2}\log\det(\mbs{\Sigma})}_{\mathcal{J}_1(\mbs{\Sigma})} + 
    \underbrace{\frac{1}{2}\sum_{i=1}^{N} \epsilon_\mathrm{i}^2}_{\mathcal{J}_2(\mbf{x}, \mbs{\Sigma})},
    \label{eq:joint_cost}
\end{equation}
where $N$ is the number of measurements. Treating the 
covariance as a fixed quantity when it's in fact 
unknown or time-varying can lead to biased state 
estimates and miscalibrated uncertainty~\cite{RN12}.

Since $\mbf{x}$ and $\mbs{\Sigma}$ are coupled 
in~\eqref{eq:joint_cost}, solving for both simultaneously in closed 
form is nontrivial. A Block-Coordinate Descent (BCD) scheme is 
therefore employed~\cite{Tseng2001BCD}, alternating between updating 
$\mbf{x}$ with $\mbs{\Sigma}$ fixed using batch 
estimation, and updating $\mbs{\Sigma}$ with 
$\mbf{x}$ fixed using the sample covariance formulation~\cite[Sec.~5.5.3]{Barfoot_2025}. Since each subproblem considered here is convex, the BCD updates satisfy the standard conditions for convergence to a stationary point~\cite{Tseng2001BCD}.

    
    

\subsection{Norm-Aware Adaptive Robust Loss Function}
To mitigate the influence of outliers, the term containing the 
Mahalanobis distance $\mathcal{J}_2(\mbf{x}, \mbs{\Sigma})$ in~\eqref{eq:joint_cost} is replaced by a 
robust loss function (RLF)~\cite{RN21},
\begin{equation}
    \mathcal{J}_2(\mbf{x},\mbs{\Sigma})
    = \sum_{\mathrm{i}=1}^N 
    \rho\!\left(\epsilon_\mathrm{i}, \alpha\right)
    = \frac{1}{2} \sum_{i=1}^N w_\mathrm{i} \, \epsilon_\mathrm{i}^2,
    \label{eq:robust_loss}
\end{equation}
where $\alpha \in (-\infty, 2]$ is a loss shape parameter controlling 
the degree of robustness. The loss function $\rho(\epsilon_\mathrm{i}, \alpha)$ 
is given by~\cite{RN21}
\begin{align}
\rho(\epsilon_\mathrm{i},\alpha)=
\begin{cases}
\displaystyle \frac{1}{2}\,\epsilon_\mathrm{i}^{2}, 
 & \alpha = 2,\\[6pt]
\displaystyle \log\!\bigl(\tfrac{1}{2}\,\epsilon_\mathrm{i}^{2}+1\bigr), 
 & \alpha = 0,\\[6pt]
\displaystyle 1-\exp\!\bigl(-\tfrac{1}{2}\,\epsilon_\mathrm{i}^{2}\bigr),
 & \alpha = -\infty,\\[8pt]
\displaystyle 
\frac{|\alpha-2|}{\alpha}
\!\left[
\Bigl(\tfrac{\epsilon_\mathrm{i}^{2}}{|\alpha-2|}+1\Bigr)^{\alpha/2}-1
\right],
 & \text{otherwise,}
\end{cases}
\label{eq:RLF}
\end{align}
where $\alpha = 2$ recovers the standard least-squares cost (L2), and 
$\alpha = 0$ recovers the Cauchy loss, which is commonly used for 
heavy-tailed outlier rejection. To establish the connection to the weighted formulation 
in~\eqref{eq:robust_loss}, the weights $w_\mathrm{i}$ are obtained 
from the loss via
\begin{equation}
    w_i = \frac{1}{\epsilon_\mathrm{i}(x_\mathrm{i})} 
    \frac{\partial \rho_\mathrm{i}(\epsilon_\mathrm{i}(x_\mathrm{i}))}
    {\partial \epsilon_\mathrm{i}(x_\mathrm{i})},
\end{equation}
yielding $w_\mathrm{i} \in [0,1]$, each depending on the current 
residual $\epsilon_\mathrm{i}$ and the optimal loss shape parameter 
$\alpha^\star$~\cite{RN20}. The weights 
$w_\mathrm{i} = w_\mathrm{i}(\epsilon_\mathrm{i}, \alpha^\star)$ 
are given by
\begin{align}
w_\mathrm{i}(\epsilon_\mathrm{i},\alpha^\star)=
\begin{cases}
\displaystyle 1,
 & \alpha^\star = 2,\\[6pt]
\displaystyle \frac{1}{\tfrac{\epsilon_\mathrm{i}^{2}}{2}+1}, 
 & \alpha^\star = 0,\\[6pt]
\displaystyle \exp \bigl(-\tfrac{1}{2}\,\epsilon_\mathrm{i}^{2}\bigr),
 & \alpha^\star = -\infty,\\[8pt]
\displaystyle 
\Bigl(\tfrac{\epsilon_\mathrm{i}^{2}}{|\alpha^\star-2|}+1\Bigr)^{{\alpha^\star/2}-1},
 & \text{otherwise,}
\end{cases}
\label{eq:weights_all}
\end{align}
where $\alpha^\star$ is found by minimizing the negative 
log-likelihood of the residual distribution~\cite{RN21},
\begin{equation}
    \alpha^\star = \arg\min_\alpha \, N \cdot \log(Z(\alpha)) + 
    \sum_{i=1}^N \rho(\epsilon_\mathrm{i}, \alpha),
    \label{eq:alpha_optimization}
\end{equation}
where $Z(\alpha) = \int_{-\infty}^{\infty} \exp(-\rho(\epsilon, \alpha))\,\mathrm{d}\epsilon$ 
is the normalization constant of the associated probability distribution. The minimization 
of~\eqref{eq:alpha_optimization} is performed using gradient descent 
with a backtracking line search~\cite{RN20}.

When the Mahalanobis distance is used as the residual, 
$\epsilon_\mathrm{i}$ follows a Chi distribution with mode 
$\tilde{\epsilon} = \sqrt{n_\mathrm{e} - 1}$~\cite[Sec.~12]{Forbes2010StatisticalDistributions}, 
where $n_\mathrm{e}$ is the error dimension. This creates a 
``mode gap'' in which inlier residuals cluster around 
$\tilde{\epsilon}$ rather than zero, causing standard RLFs to 
inadvertently downweight inliers~\cite{RN20}.

To address this, a Maxwell-Boltzmann distribution is fit to the 
residuals to robustly estimate $\tilde{\epsilon}$~\cite{RN20}, and 
mode-shifted residuals $\xi_\mathrm{i} = \epsilon_\mathrm{i} - \tilde{\epsilon}$ are introduced for all $\epsilon_\mathrm{i} \geq 
\tilde{\epsilon}$. The adaptive loss 
from~\eqref{eq:alpha_optimization} is then applied only to the 
shifted residuals, while all residuals below the mode are treated as 
inliers and assigned a weight of one. The final norm-aware weights 
are

\begin{equation}
\tilde{w}_\mathrm{i}(\epsilon_\mathrm{i}, \alpha^\star)
=
\begin{cases}
1, & \text{if } \epsilon_\mathrm{i} < \tilde{\epsilon}, \\[6pt]
w_\mathrm{i}(\xi_\mathrm{i}, \alpha^\star), & \text{otherwise},
\end{cases}
\label{eq:norm_aware_weights}
\end{equation}
where $w_\mathrm{i}(\xi_\mathrm{i}, \alpha^\star)$ is given 
by~\eqref{eq:weights_all}.

The norm-aware weights~\eqref{eq:norm_aware_weights} are incorporated 
into the batch estimator through an IRLS scheme~\cite{HollandWelsch1977}, which 
alternates between computing the weights and updating the state. 
Given the current weights, the state is refined via a weighted 
Gauss-Newton step~\cite[Sec.~4.3]{Barfoot_2025}, where the weight 
matrix becomes
\begin{equation}
\begin{split}
    \mbf{W} = \mathrm{blockdiag}\!\left(\,\check{\mbf{P}}_0^{-1}, 
    \mbf{Q}_\mathrm{0}^{-1}, \ldots, \mbf{Q}_\mathrm{M-1}^{-1},\right. \\
    \left.\tilde{w}_1 \mbs{\Sigma}^{-1}, \ldots,
    \tilde{w}_\mathrm{N} \mbs{\Sigma}^{-1}\right),
\end{split}
\end{equation}
combining the robust weights $\tilde{w}_\mathrm{i}$ with the inverse 
measurement covariance $\mbs{\Sigma}^{-1}$, where $\check{\mbf{P}}_0$ 
is the prior state covariance, $\mbf{Q}_\mathrm{j}$ is the process 
noise covariance at time step $\mathrm{j}$ for $M$ process steps, and $N$ 
is the number of exteroceptive measurements, all assumed to be known. 
This process is repeated until convergence of the state estimates.

\subsection{Minimum Weighted Covariance Determinant}
The sample covariance,
\begin{equation}
    \hat{\mbs{\Sigma}}_{\mathrm{ML}}
    = \frac{1}{N} \sum_{i=1}^{N} \mbf{e}_\mathrm{i} \mbf{e}_\mathrm{i}^\trans,
    \label{eq:sample_cov}
\end{equation}
is highly sensitive to outliers, as a single 
large residual can severely bias the covariance estimate. While incorporating 
residual weights into the covariance estimate~\cite[Sec.~5.5.1]{Barfoot_2025} 
reduces the influence of extreme residuals, the estimate can still 
be biased when outlier clusters are present, as these induce a 
multimodal residual distribution that a unimodal weighted estimator 
cannot adequately capture. A more robust covariance estimator is therefore required.

The Minimum Covariance Determinant (MCD) estimator~\cite{RN22} 
identifies the subset $\mathcal{H}^\star \subset \{1, \ldots, N\}$ 
of size $h$, where $|\mathcal{H}| = h$ denotes the number of 
elements in the subset, whose empirical covariance has the smallest determinant,
\begin{equation}
    \mathcal{H}^\star
    = \arg\min_{|\mathcal{H}| = h}
    \det\!\left(
        \sum_{i \in \mathcal{H}}
            \mbf{e}_\mathrm{i} \mbf{e}_\mathrm{i}^\trans
    \right),
\end{equation}
which offers a breakdown point of up to 50\%~\cite{RN22}. A 
reweighting step is then applied, assigning binary weights 
$b_\mathrm{i} \in \{0, 1\}$ based on the Mahalanobis distance 
to the MCD estimate, with the decision rule $b_\mathrm{i} = 0$ if 
$\epsilon_\mathrm{i}^{2} > \chi^2_{\mathrm{n_e},\,0.999}$ and $b_\mathrm{i} = 1$ 
otherwise. Any measurement whose squared Mahalanobis distance is statistically 
incompatible with the assumed noise model at the 99.9\% confidence 
level is excluded, and the covariance is recomputed over the remaining subset. 
The support 
$\mathcal{H}^\star$ obtained from this step is then used to reestimate the covariance using the continuous residual weights 
$\tilde{w}_\mathrm{i}$ from~\eqref{eq:norm_aware_weights}, yielding the Minimum 
Weighted Covariance Determinant (MWCD) 
estimator~\cite{Kalina2022WeightedMCD},
\begin{equation}
    \hat{\mbs{\Sigma}}_{\mathrm{MWCD}}
    =
    \frac{1}{\sum_{i \in \mathcal{H}^\star} \tilde{w}_\mathrm{i}}
    \sum_{i \in \mathcal{H}^\star}
        \tilde{w}_\mathrm{i}\, \mbf{e}_\mathrm{i} \mbf{e}_\mathrm{i}^\trans.
    \label{eq:mwcd_covariance}
\end{equation}
This formulation preserves the high breakdown point of the MCD 
while leveraging residual weights that vary continuously with the 
residual magnitude, rather than applying a binary decision, 
allowing the estimator to adapt to different outlier profiles and 
yielding a more accurate covariance reestimation under heavy-tailed 
and outlier-contaminated measurements.

    \section{Method Summary}
\label{sec:adaptive_robust}

This section presents the proposed adaptive robust joint state and 
covariance estimation framework, which combines the components 
introduced in Section~\ref{sec:preliminaries} within a unified BCD 
procedure.

\subsection{Adaptive Robust Joint State and Covariance Estimation}
The proposed framework jointly estimates the system state 
$\hat{\mbf{x}}$, the measurement covariance $\hat{\mbs{\Sigma}}$, 
and the loss shape parameter $\alpha$ through a BCD procedure that 
alternates between three updates at each iteration. The estimator 
is initialized with a positive definite covariance 
$\hat{\mbs{\Sigma}}^{(0)}$ and a loss shape parameter $\alpha^{(0)} \in 
(0, 1]$, where values in this range are empirically found to yield 
faster convergence on a problem-by-problem basis.

At each iteration, the state is refined through an IRLS 
step~\cite{HollandWelsch1977}~\cite[Sec.~9]{Boyd_2004}, which internally 
computes the norm-aware weights $\tilde{w}_\mathrm{i}$ 
via~\eqref{eq:norm_aware_weights}.

Given the updated state, the measurement covariance is reestimated 
via the MWCD estimator~\eqref{eq:mwcd_covariance} using the weights 
$\tilde{w}_\mathrm{i}$ previously found using~\eqref{eq:norm_aware_weights}. A key feature of the proposed framework is 
that both the IRLS state update and the MWCD covariance update rely 
on the same norm-aware weight function~\eqref{eq:norm_aware_weights}, 
ensuring that the treatment of outliers is consistent across the 
state and covariance estimation steps.

Finally, the loss shape parameter $\alpha$ is updated by 
solving~\eqref{eq:alpha_optimization}, enabling the loss to 
automatically adapt its robustness level to the empirical residual 
distribution. These three updates are repeated until the convergence 
criteria are met.

The Wasserstein-2 distance between two zero-mean Gaussian distributions 
with covariances $\mbs{\Sigma}_1$ and $\mbs{\Sigma}_2$ is given 
by~\cite[Sec.~2.6]{peyré2020computationaloptimaltransport},
\begin{equation}
    W_2(\mbs{\Sigma}_1, \mbs{\Sigma}_2) = 
    \sqrt{\mathrm{tr}\!\left(
        \mbs{\Sigma}_1 + \mbs{\Sigma}_2
        - 2\left(
            \mbs{\Sigma}_2^{1/2} 
            \mbs{\Sigma}_1
            \mbs{\Sigma}_2^{1/2}
        \right)^{\!1/2}
    \right)}.
    \label{eq:w2}
\end{equation}
The full procedure is summarized in Algorithm~\ref{alg:adaptive_robust}, 
where convergence is declared when $\|\delta\mbf{x}\|_2 < \tau_x$, 
$|\nabla_\alpha \mathcal{J}| < \tau_\alpha$, where $\nabla_\alpha 
\mathcal{J}$ is the gradient of the cost with respect to the shape 
parameter, and $W_2(\hat{\mbs{\Sigma}}^{(t+1)}, \hat{\mbs{\Sigma}}^{(t)}) < 
\tau_\Sigma$ as defined in~\eqref{eq:w2}. The tolerances $\tau_x$, 
$\tau_\alpha$, and $\tau_\Sigma$ are user-defined tolerances.

\begin{algorithm}
\caption{Adaptive Robust Joint State and Covariance Estimation}
\label{alg:adaptive_robust}
\begin{algorithmic}[1]
    \State \textbf{Initialize} $\hat{\mbs{\Sigma}}^{(0)}$ and 
    $\alpha^{(0)} \in (0, 1]$
    \Repeat
        \Repeat
            \State Compute $\tilde{w}_\mathrm{i}$ 
            via~\eqref{eq:norm_aware_weights}
            \State Update $\hat{\mbf{x}}$ via Gauss-Newton
        \Until{$\|\delta\mbf{x}\|_2 < \tau_x$}
        \State Update $\hat{\mbs{\Sigma}}$ via 
        MWCD~\eqref{eq:mwcd_covariance}
        \State Update $\alpha^\star$ 
        via~\eqref{eq:alpha_optimization}
    \Until{$|\nabla_\alpha \mathcal{J}| < \tau_\alpha$ and 
    $W_2(\hat{\mbs{\Sigma}}^{(t+1)}, \hat{\mbs{\Sigma}}^{(t)}) 
    < \tau_\Sigma$}
\end{algorithmic}
\end{algorithm}
    \section{Results}
\label{sec:results}

The proposed framework is evaluated in two settings, a Monte-Carlo 
simulation and a real-world UWB localization experiment, both using a 
random walk process model. All computations are implemented using 
\texttt{navlie}, a Python package for state estimation~\cite{Navlie}. 
In this experiment, the UWB range is one-dimensional, so the error 
dimension is $n_e = 1$ and the Chi-distribution mode 
$\tilde{\epsilon} = \sqrt{n_e - 1}$ is zero. However, the 
proposed pipeline remains unchanged, 
as a mode of zero is naturally accommodated within 
the problem formulation.

The goal in both simulation and experimental settings is to estimate the robot position 
$\mbf{x}_\mathrm{i} = (x_\mathrm{i}, y_\mathrm{i})$ at each time step $\mathrm{i}$. 
To assess the performance of the 
proposed approach in a vehicle-agnostic way, no informative process model 
is exploited. Rather, a random walk process model is used,
\begin{equation}
    \mbf{x}_\mathrm{i} = \mbf{x}_\mathrm{i-1} + \mbf{v}_\mathrm{i},
    \qquad
    \mbf{v}_\mathrm{i} \sim \mathcal{N}(\mbf{0}, \mbf{Q}),
    \label{eq:process_model}
\end{equation}
where $\mbf{v}_\mathrm{i}$ is the process noise. The measurement 
model is given by
\begin{equation}
    y_\mathrm{i} = \left\Vert \mbf{x}_\mathrm{i} - \mbf{a}_\mathrm{i} \right\Vert_2 + e_\mathrm{i},
    \qquad
    e_\mathrm{i} \sim \mathcal{N}(0, R),
    \label{eq:range_model}
\end{equation}
where $y_\mathrm{i}$ is the measured range, $\mbf{a}_\mathrm{i}$ is 
the position of the anchor associated with the $\mathrm{i}$-th 
measurement, and $e_\mathrm{i}$ is the measurement noise with 
covariance $R$.

Five joint state and covariance estimators are evaluated, with the 
loss shape parameter $\alpha$ as defined in~\eqref{eq:RLF}. The L2 
estimator is the standard joint state and covariance estimator with 
no outlier robustness~\cite{RN12}. The Charbonnier ($\alpha = 1$) and 
Cauchy ($\alpha = 0$) estimators are fixed-loss robust joint state 
and covariance estimators. These values, $\alpha = 2$, $\alpha = 1$, 
and $\alpha = 0$, were selected through manual tuning, spanning the 
range of loss functions, and represent 
the regime a practitioner would naturally explore for heavy-tailed 
positively skewed outliers of this type. The MCD estimator is used in an 
ablation study of the proposed framework where in the MWCD covariance 
update is replaced by the plain MCD estimator, isolating the 
contribution of the continuous norm-aware weights in the covariance 
reestimation. The proposed adaptive robust estimator adapts both the 
loss shape parameter and the measurement covariance sequentially, as 
described in Section~\ref{sec:adaptive_robust}.

Performance is assessed using two metrics. The position estimation 
accuracy is measured by the Root Mean Square Error (RMSE), computed 
against the ground truth as
\begin{equation}
    \mathrm{RMSE} = \sqrt{\frac{1}{N}\sum_{i=1}^{N} \|\hat{\mbf{x}}_\mathrm{i} - \mbf{x}_{\mathrm{i}, {\mathrm{gt}}}\|^2_2},
    \label{eq:rmse}
\end{equation}
where $\mbf{x}_{\mathrm{i}, {\mathrm{gt}}}$ is the ground truth state. 
The covariance estimation quality is measured by the Wasserstein-2 
distance between the estimated and true measurement 
covariance in~\eqref{eq:w2}.

\subsection{Monte-Carlo Simulation}
A Monte-Carlo simulation is conducted to evaluate the estimators 
under controlled inlier and outlier conditions. A robot with 
a single UWB tag moves while receiving range measurements 
to nine UWB anchors. Inlier measurement noise is drawn 
from a zero-mean Gaussian with covariance $R = \sigma^2$, and a controlled 
fraction of measurements are corrupted with positive biases drawn from 
a lognormal distribution with a mean of $2.5\sigma$, where $\sigma$ is 
the inlier standard deviation, emulating the positively skewed errors 
characteristic of NLOS propagation. The outlier fraction is varied 
from \SI{0}{\percent} to \SI{50}{\percent}, and each condition is 
averaged over $30$ Monte-Carlo trials.

\begin{figure}[htbp!]
    \centering
    \includegraphics[width=\columnwidth]{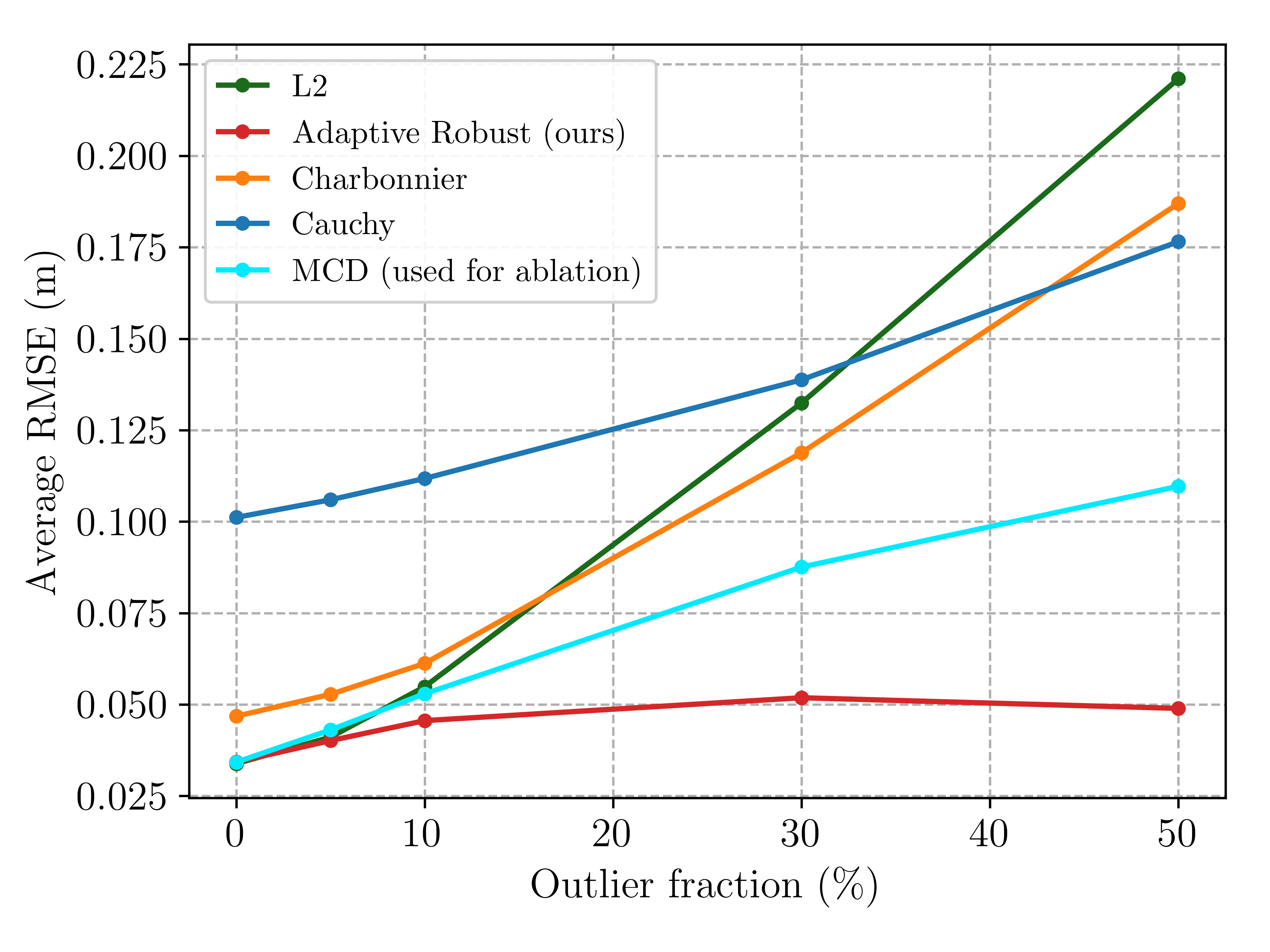}
    \caption{Position RMSE as a function of the outlier fraction, 
    averaged over $30$ Monte-Carlo trials.}
    \label{fig:rmse_simulation}
\end{figure}
\FloatBarrier

\begin{figure}[htbp!]
    \centering
    \includegraphics[width=\columnwidth]{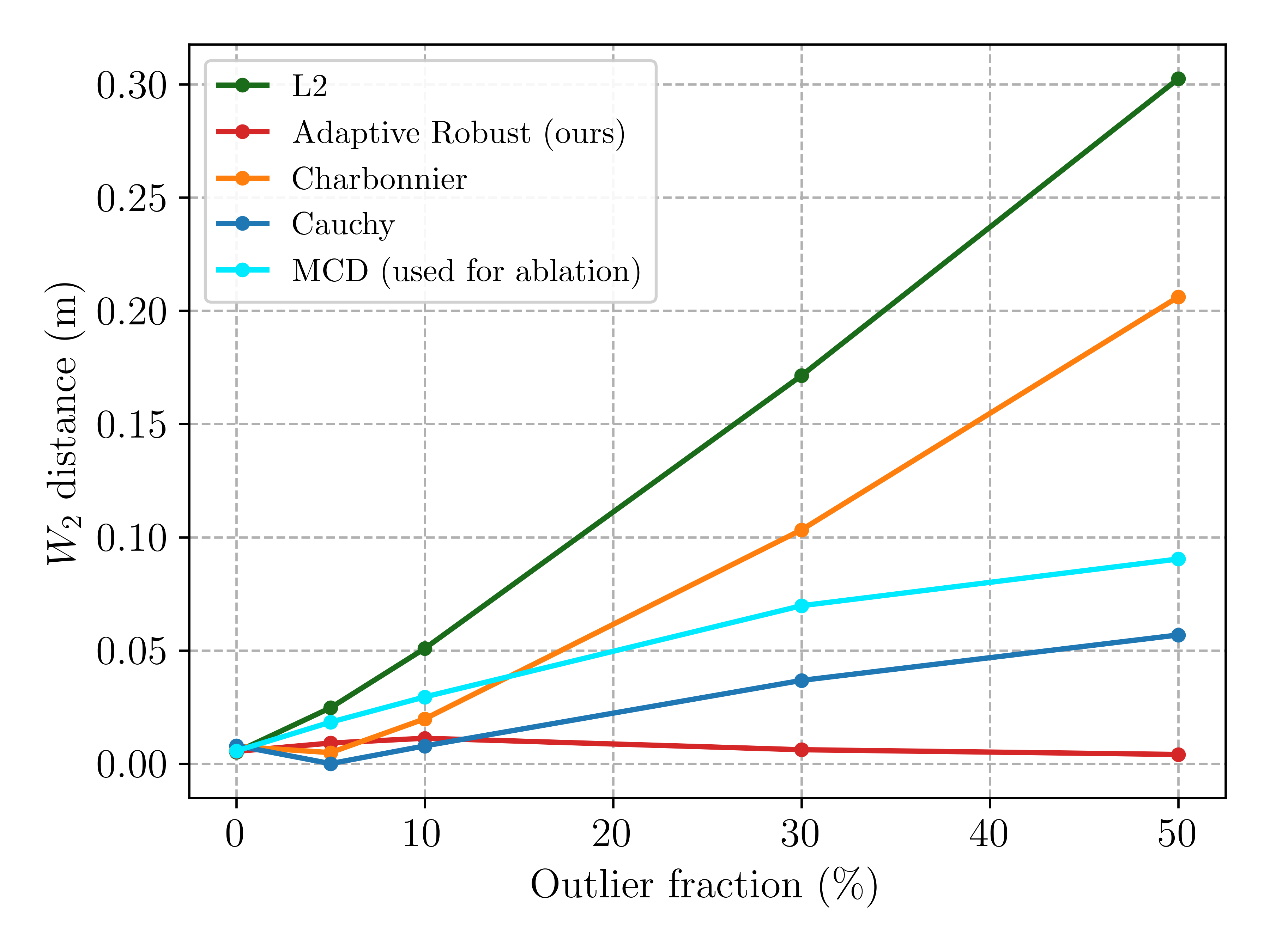}
    \caption{Wasserstein-2 distance between the estimated and true 
    measurement covariance as a function of the outlier fraction, 
    averaged over $30$ Monte-Carlo trials.}
    \label{fig:r_error_simulation}
\end{figure}
\FloatBarrier

Fig.~\ref{fig:rmse_simulation} reports the position RMSE and 
Fig.~\ref{fig:r_error_simulation} the $W_2$ distance between the 
estimated and true measurement covariance, both as a function of 
the outlier fraction. The adaptive robust estimator (ours) maintains a 
consistently low RMSE and $W_2$ distance across all outlier 
fractions, with both remaining nearly flat even as the contamination 
reaches \SI{50}{\percent}. In contrast, the L2 estimator degrades 
steadily in both RMSE and $W_2$ distance as outliers are introduced, 
since it assigns equal weight to all measurements. Charbonnier 
achieves an RMSE and $W_2$ distance close to the adaptive robust 
estimator at low outlier fractions, but both grow progressively as 
contamination increases, reflecting the limitation of a fixed loss 
shape. Cauchy maintains a low $W_2$ distance but exhibits a higher 
RMSE that grows with the outlier fraction, as its aggressive 
downweighting excludes inliers alongside outliers. The MCD ablation 
keeps a lower RMSE and $W_2$ distance than the fixed-loss baselines, 
but both remain higher than those of the full framework, confirming 
that the continuous norm-aware weights of the MWCD update improve 
both covariance recovery and state estimation as contamination grows.

\subsection{Real-World Experiment}
\begin{figure}[htbp!]
    \centering
    \includegraphics[width=\columnwidth]{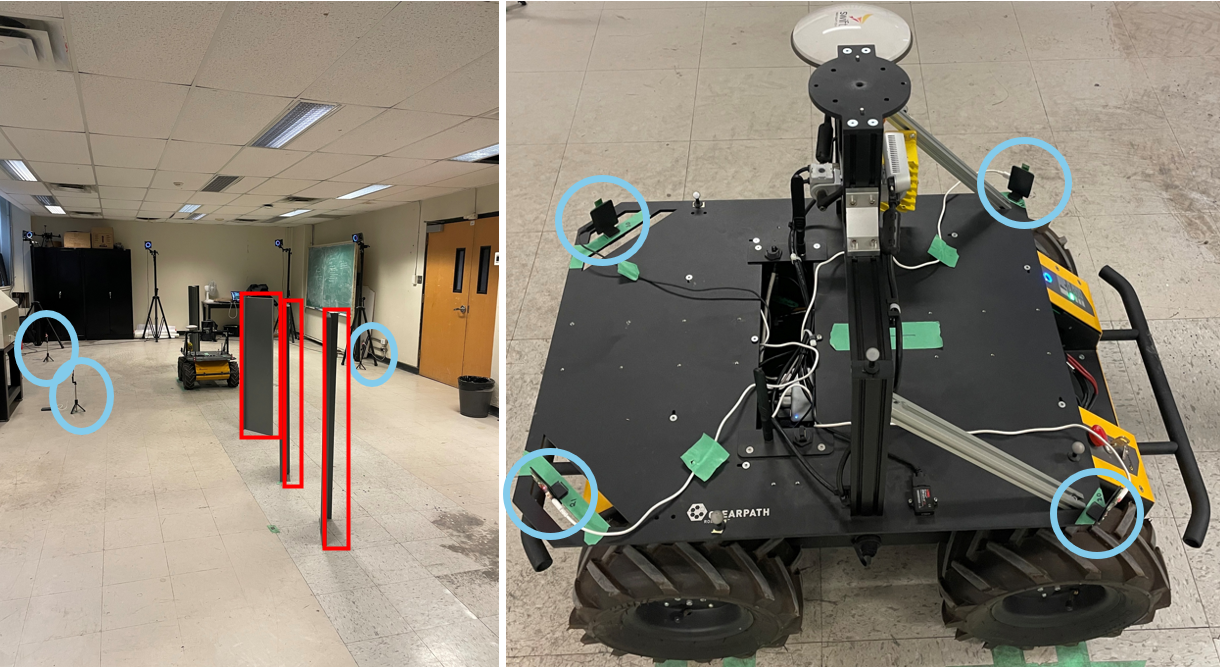}
    \caption{The Clearpath Husky unmanned ground vehicle navigating a cluttered 
    environment. UWB tags mounted on the robot and fixed anchors are 
    circled in blue. Metal obstacles placed to induce NLOS conditions 
    are outlined in red.}
    \label{fig:experiment_image}
\end{figure}

A Clearpath Husky unmanned ground vehicle equipped with four 
Decawave DWM1000 UWB tags was used for data collection, as shown 
in Fig.~\ref{fig:experiment_image}. Each tag provides range 
measurements to a set of fixed UWB anchors placed throughout the 
test environment. Because UWB ranging estimates distance from 
the time-of-flight of radio signals, it is sensitive to NLOS 
propagation, which induces positive range biases and 
heavy-tailed errors~\cite{Guvenc2009NLOS}. Further degradation 
arises from antenna delays, inter-tag clock skews, and 
non-uniform radiation patterns~\cite{10160769}.

\begin{figure}[htbp!]
    \centering
    \includegraphics[width=\columnwidth]{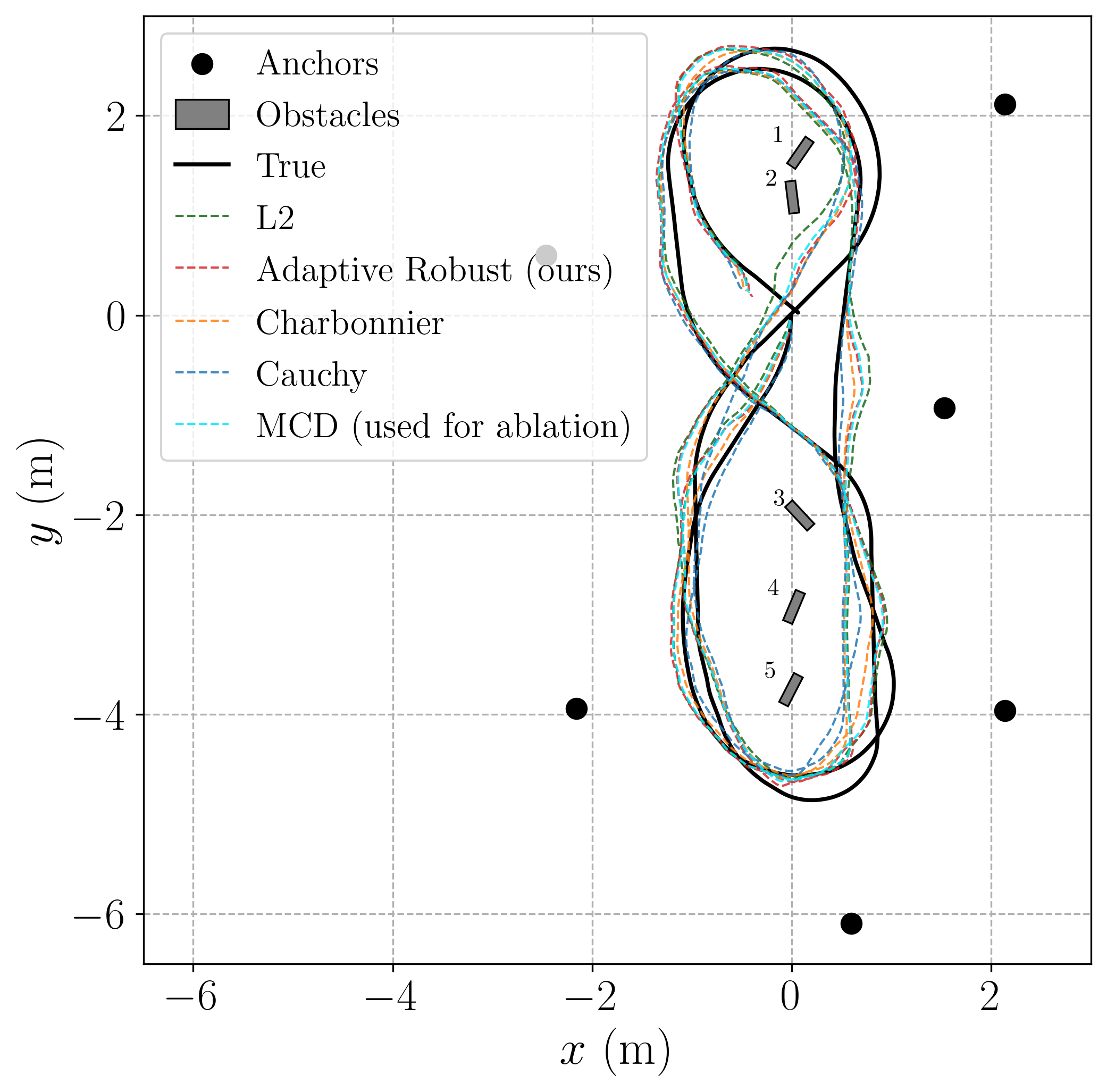}
    \caption{Trajectory 5 of the Husky in the cluttered environment using 
    the random walk process model. Estimated paths from all five 
    estimators are shown against the motion-capture ground truth. Black 
    dots denote UWB anchor locations and grey rectangles denote metal 
    obstacles.}
    \label{fig:trajectory_plot}
\end{figure}

The Husky was driven along ten independent trajectories within a 
\SI{40}{\metre\squared} indoor area at a nominal forward velocity 
of \SI{0.4}{\metre\per\second}, yielding trajectories of 
approximately \SI{90}{\second} each. Four trajectories were 
collected in an uncluttered environment with no obstacles. The remaining six trajectories were collected with metal obstacles 
placed in the center of the room to induce NLOS conditions. 
Trajectories 1 and 2 include obstacles 1, 3, and 5; trajectories 
3 and 4 add obstacle 2; and trajectories 5 and 6 further add 
obstacle 4, as visible in 
Fig.~\ref{fig:trajectory_plot}. UWB range measurements were 
collected at \SI{30}{\hertz} per tag, and ground-truth poses were 
recorded at \SI{120}{\hertz} using a motion capture system. 
The UWB modules were calibrated for 
antenna delays and pose-dependent biases following the procedure 
of~\cite{10160769}, in order to isolate the effects of NLOS 
propagation. The state is initialized via 
multilateration using a batch of UWB measurements, 
providing an initial trajectory estimate~\cite{FOY}.

\begin{figure}[htbp!]
    \centering
    \begin{subfigure}[t]{\columnwidth}
        \centering
        \includegraphics[width=\textwidth]{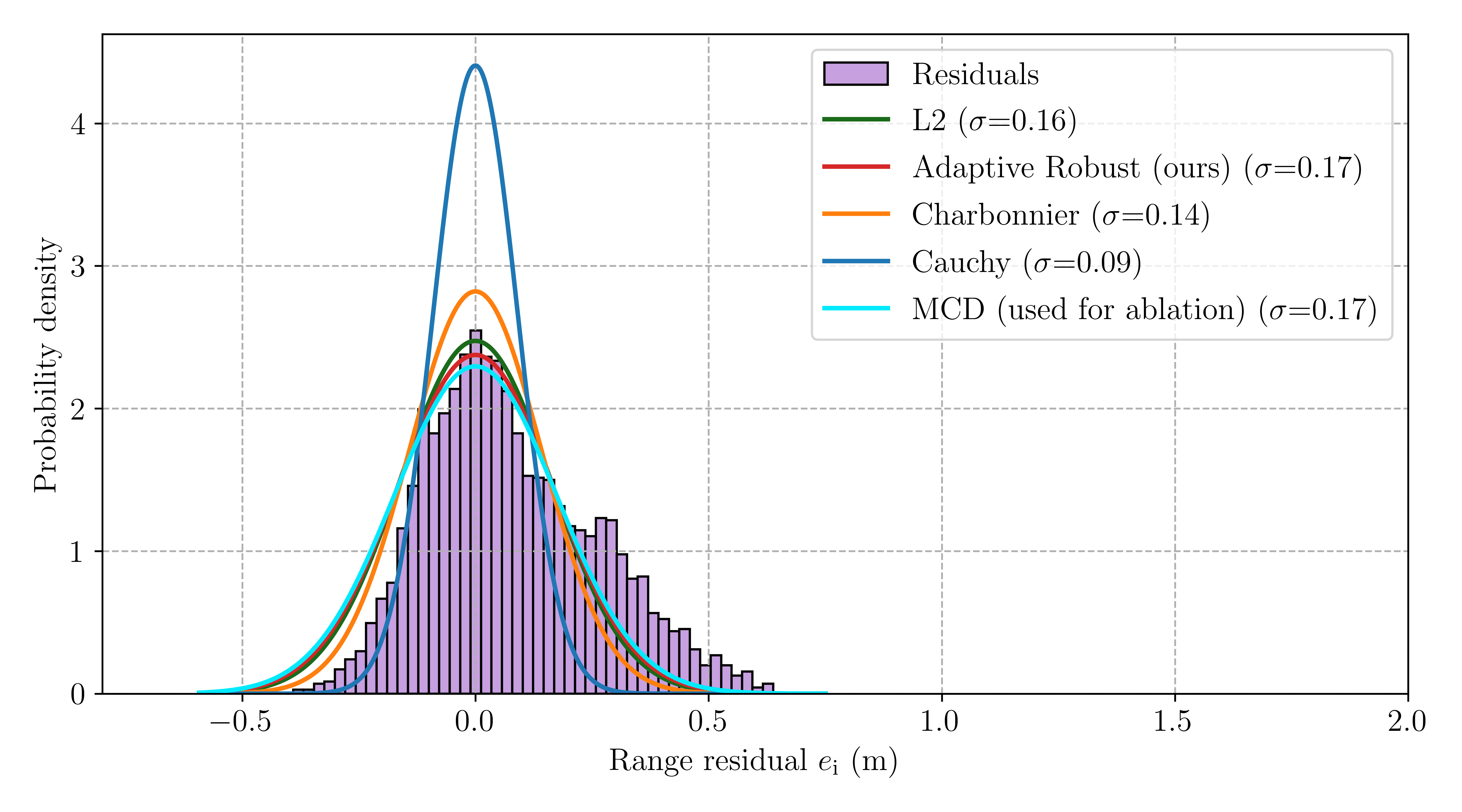}
        \caption{Uncluttered environment.}
        \label{fig:hist_uncluttered_ml}
    \end{subfigure}
    \vspace{0.5em}
    \begin{subfigure}[t]{\columnwidth}
        \centering
        \includegraphics[width=\textwidth]{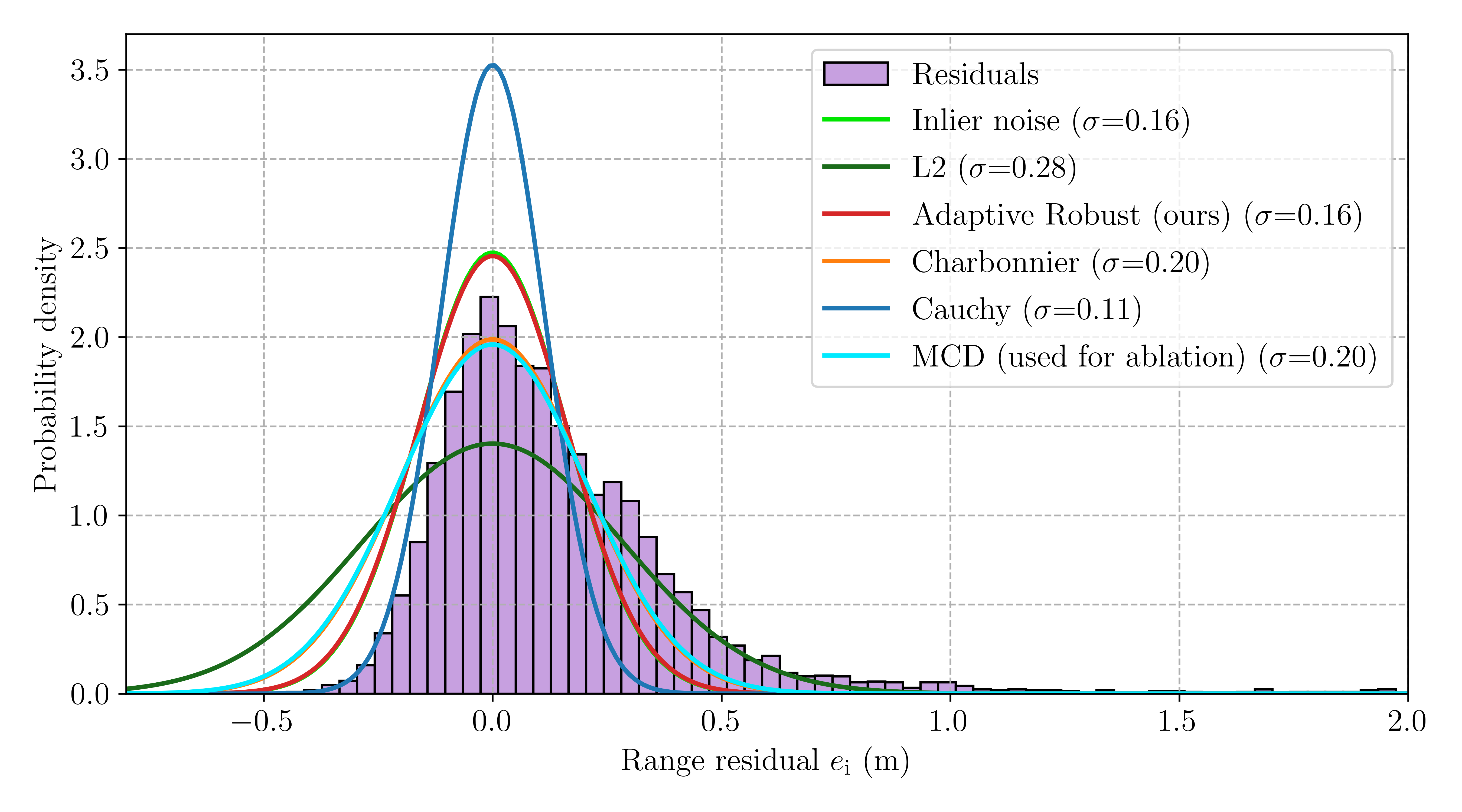}
        \caption{Cluttered environment.}
        \label{fig:hist_cluttered_ml}
    \end{subfigure}
    \caption{Range residual distributions and fitted covariance estimates for each estimator in the random walk experiment.}
    \label{fig:histograms_ml}
\end{figure}

For the uncluttered trajectories, an MCD 
estimator is first applied to the residuals between the measured 
UWB ranges and the ground-truth ranges to obtain an initial 
covariance estimate, after which measurements lying outside the 
$3\sigma$, where $\sigma$ is the standard deviation of the 
estimated covariance, bounds are removed to obtain an approximately Gaussian 
residual distribution. The L2 covariance estimate is then computed 
for each uncluttered trajectory and averaged across trajectories, 
yielding a reference inlier noise level against which the 
covariance estimates in the cluttered environment are compared. 
As confirmed by the uncluttered histograms in 
Fig.~\ref{fig:histograms_ml}, the L2 estimator yields a 
close fit to the residual distribution in the absence of outliers, 
validating its use as a reference.

Fig.~\ref{fig:histograms_ml} shows the range residual distributions 
and fitted covariance estimates for all estimators in both 
environments. In the uncluttered environment, the residuals are 
approximately Gaussian and the adaptive robust estimator recovers 
a standard deviation ($\sigma = 0.17$\,m) close to the L2 estimate 
($\sigma = 0.16$\,m), which serves as the reference in the absence 
of outliers. In the cluttered environment, the residual distribution 
exhibits the characteristic heavy-tailed positively skewed profile 
induced by NLOS propagation. The adaptive robust estimator (ours) recovers 
a standard deviation ($\sigma = 0.16$\,m) essentially identical to the true 
inlier noise level ($\sigma = 0.16$\,m), while the L2 estimator 
significantly overestimates ($\sigma = 0.28$\,m) due to its 
sensitivity to outlier-corrupted residuals, Charbonnier 
overestimates moderately ($\sigma = 0.20$\,m), and Cauchy 
underestimates ($\sigma = 0.11$\,m) by over-suppressing residuals. The MCD 
ablation also overestimates ($\sigma = 0.20$\,m), showing 
that replacing the continuous norm-aware weights with the plain 
MCD reestimation degrades covariance recovery despite an otherwise 
identical pipeline.

\FloatBarrier

\begin{table}[htbp!]
\caption{Position RMSE (m) for the random walk experiment.}
\label{tab:rmse_ml}
\centering
\resizebox{\columnwidth}{!}{%
\begin{tabular}{l|c|cccccc c}
\toprule
 & {Uncluttered} & \multicolumn{7}{c}{Cluttered} \\
\cmidrule(lr){2-2} \cmidrule(lr){3-9}
Method & Avg. & 1 & 2 & 3 & 4 & 5 & 6 & Avg. \\
\midrule
L2          & \textbf{0.081} & 0.117 & 0.115 & 0.114 & 0.114 & 0.135 & 0.112 & 0.118 \\
Adaptive (ours)    & \textbf{0.081} & \textbf{0.091} & \textbf{0.089} & \textbf{0.084} & 0.101 & 0.117 & 0.099 & \textbf{0.097} \\
Charbonnier & 0.085 & 0.095 & 0.091 & 0.093 & \textbf{0.096} & \textbf{0.115} & \textbf{0.095} & 0.098 \\
Cauchy      & 0.107 & 0.097 & 0.117 & 0.122 & \textbf{0.096} & 0.118 & \textbf{0.095} & 0.107 \\
MCD (used for ablation)         & 0.082 & 0.095 & 0.095 & 0.094 & 0.103 & 0.121 & 0.100 & 0.101 \\
\bottomrule
\end{tabular}%
}
\end{table}
\FloatBarrier

Table~\ref{tab:rmse_ml} reports the position RMSE across all 
trajectories for both environments. In the uncluttered environment, 
the adaptive robust estimator achieves the same average RMSE as L2 
($0.081$\,m), confirming that the added robustness mechanism does 
not degrade performance in the absence of outliers. Cauchy exhibits 
a notably higher average RMSE ($0.107$\,m) even in the uncluttered 
case, reflecting the cost of its aggressive downweighting of 
residuals. In the cluttered environment, the adaptive robust 
estimator achieves the lowest average RMSE of $0.097$\,m, 
outperforming L2 ($0.118$\,m), Charbonnier ($0.098$\,m), and 
Cauchy ($0.107$\,m), demonstrating its ability to suppress 
NLOS-induced outliers while maintaining accurate state estimation. 
The MCD ablation achieves a higher average RMSE ($0.101$\,m) than 
the full framework, indicating that the continuous norm-aware 
weights in the MWCD update also benefit state estimation accuracy.
    \section{Discussion}
\label{sec:discussion}

In the weighted least-squares formulation, the state update 
balances contributions from the process model and the measurements 
through the weight matrix $\mbf{W}$. A poorly estimated measurement covariance directly 
distorts this balance. When $\hat{\mbs{\Sigma}}$ is overestimated, 
as with L2 which assigns unit weights to all measurements, 
too many outlier-corrupted measurements are included in the state 
update, biasing the estimate. When $\hat{\mbs{\Sigma}}$ is 
underestimated, as with Cauchy, the inflated Mahalanobis distances 
drive the robust weights toward zero for both inliers and outliers, 
effectively excluding most measurements from the state update and 
shifting reliance toward the process model. This 
underestimation is visible in Fig.~\ref{fig:histograms_ml}, where the Cauchy fitted 
distribution is too narrow to capture the inlier residual spread.

\begin{figure}[htbp!]
    \centering
    \includegraphics[width=\columnwidth]{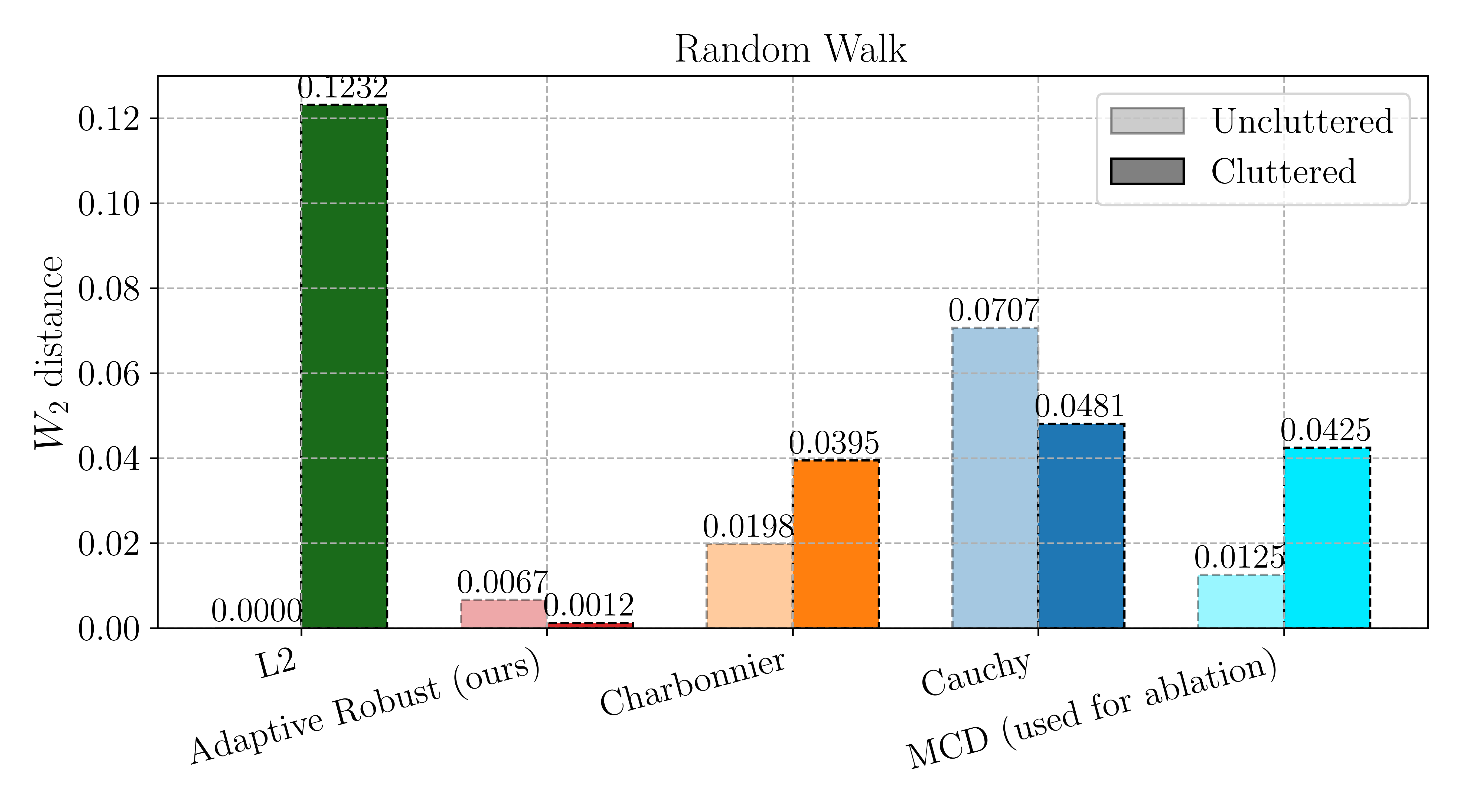}
    \caption{Wasserstein-2 distance between the estimated and true 
    measurement covariance for each estimator across LOS and NLOS experiments. 
    The L2 uncluttered $W_2$ distance is zero as it 
    serves as the reference inlier noise level.}
    \label{fig:w2_barplot}
\end{figure}
\FloatBarrier

This mechanism explains why the fixed-loss baselines degrade as the 
outlier fraction grows in the simulation, while the adaptive robust 
estimator remains stable. Each fixed loss commits to a single 
weighting behavior, so it is only well matched to a narrow range of 
contamination, whereas adapting the loss shape parameter and the 
recovered covariance to the empirical residual distribution keeps 
the weighting appropriate across all conditions.

The same effect appears in the real-world experiment. Charbonnier, 
with $\alpha = 1$, achieves an RMSE similar to the adaptive robust 
estimator, as it represents the loss a practitioner familiar with 
this outlier profile would likely select through manual tuning. 
However, it overestimates the inlier covariance in the cluttered 
environment, as visible in Fig.~\ref{fig:histograms_ml}, resulting 
in a miscalibrated uncertainty estimate, and requires manual 
parameter tuning. The adaptive robust estimator recovers the true 
inlier covariance accurately without any parameter tuning, yielding 
a well-calibrated uncertainty alongside competitive state estimation 
accuracy, as confirmed by the $W_2$ distances in 
Fig.~\ref{fig:w2_barplot}.

The MCD ablation isolates the role of the continuous weighting in 
the covariance update. While the MCD subset selection rejects the 
bulk of the outlier cluster, its binary inclusion rule weights all 
retained measurements equally, leaving the estimate biased by 
partially corrupted residuals. The continuous 
norm-aware weights of the MWCD down-weight these residuals according 
to their magnitude, recovering a covariance closer to the true inlier 
noise level.

This covariance recovery is confirmed by the $W_2$ distances shown 
in Fig.~\ref{fig:w2_barplot}, where the adaptive robust estimator 
achieves the lowest $W_2$ distance across both LOS and NLOS experiments. For 
the adaptive robust estimator, the uncluttered $W_2$ distance is 
marginally higher than the cluttered one, though the two remain 
nearly identical. Across all conditions, the estimated covariance 
remains close to the estimated true inlier noise level, 
demonstrating that the proposed MWCD-based covariance update 
reliably identifies and rejects outlier-corrupted measurements 
while preserving the statistical integrity of the inlier subset.

    \section{Conclusions}
\label{sec:conclusion}

The contribution of this paper is an adaptive robust framework 
for joint state and covariance estimation under outlier-contaminated 
measurements. The proposed method combines a norm-aware adaptive 
robust loss, an IRLS state update, and a MWCD covariance estimator 
within a unified BCD procedure that jointly adapts the state 
estimate, the measurement covariance, and the loss shape parameter 
without any manual tuning.

The framework was validated in a Monte-Carlo simulation and on 
real-world UWB localization experiments using a Husky unmanned 
ground vehicle navigating cluttered NLOS environments. Across both 
settings, the adaptive robust estimator consistently recovered a 
measurement covariance close to the inlier noise level, 
achieving the lowest $W_2$ distance among all estimators. In the 
simulation, it maintained a low RMSE and $W_2$ distance across 
outlier fractions up to \SI{50}{\percent}, while the fixed-loss 
baselines degraded as outlier contamination increased. In the real-world 
experiment, it outperformed all baselines in state estimation 
accuracy, though Charbonnier, the best manually tuned fixed-loss 
alternative for this outlier profile, achieved a comparable RMSE, 
but required manual tuning and produced a miscalibrated uncertainty 
estimate.

These results demonstrate that accurately recovering the inlier 
measurement covariance is critical for reliable state estimation, 
as it ensures that valid measurements are properly incorporated 
into the state update. The proposed framework consistently delivers 
accurate covariance estimates and better state estimation accuracy 
across varying environments, establishing it as a robust and 
self-tuning solution for real-world localization under 
NLOS-contaminated measurements.
    \section*{Acknowledgment}
The authors acknowledge the use of Microsoft Copilot to review an initial draft of this 
paper. The authors reviewed and edited the content of the paper as needed 
and take full responsibility for the content of this paper.
    \printbibliography
\end{document}